\documentclass[]{TEAI}

\usepackage{helvet}
\usepackage[utf8]{inputenc}
\usepackage[T1]{fontenc}

\usepackage{amsmath,amssymb,mathtools,amsfonts}
\usepackage{graphicx}
\usepackage{subcaption}
\usepackage{wrapfig}
\usepackage{caption}

\usepackage{booktabs}
\usepackage{multirow}
\usepackage{enumitem}
\usepackage{xspace}
\usepackage{tabularx}
\usepackage{nicefrac}
\usepackage{array}

\usepackage{xcolor}
\usepackage{hyperref}
\usepackage{url}

\usepackage{natbib}

\usepackage{placeins}
\usepackage{float}
\usepackage{tikz}
\usetikzlibrary{arrows.meta,positioning}

\usepackage{tcolorbox}
\tcbuselibrary{breakable,skins}
\usepackage{fvextra}
\fvset{breaklines=true,breakanywhere=true,fontsize=\scriptsize,breaksymbolleft={},breaksymbolsepleft=0pt}

\definecolor{bestcolor}{HTML}{E8F4EA}
\definecolor{highlight}{HTML}{F4F4F4}
\definecolor{gaincolor}{HTML}{1B7A2C}
\definecolor{losscolor}{HTML}{B11A2A}

\usepackage{catchfile}

\providecommand{\email}[1]{\href{mailto:#1}{#1}}

\newcommand{\method}{\textsc{TraceGraph}\xspace}
\newcommand{\sg}{\textsc{SliceGraph}\xspace}
\newcommand{\mcp}{\textsc{MCPBench}\xspace}
\newcommand{\swe}{\textsc{SWE-bench}\xspace}
\DeclareRobustCommand{\taub}{\texorpdfstring{\ensuremath{\tau^{2}}\textsc{-bench}}{tau2-bench}}
\newcommand{\terminal}{\textsc{TerminalBench}\xspace}
\newcommand{\searchb}{\textsc{Search}\xspace}
\newcommand{\idf}{\mathrm{idf}}

\newcommand{\E}{\mathbb{E}}

\title{\method{}: Shared Decision Landscapes for Diagnosing and Improving Agent Trajectories}

\author{
Junjie Nian\textsuperscript{1*},
Kang Chen\textsuperscript{1*},
Ge Zhang\textsuperscript{2},
Yixin Cao\textsuperscript{1,3$\dagger$},
Yugang Jiang\textsuperscript{1}
}

\affiliation[1]{\mbox{Fudan University}}
\affiliation[2]{\mbox{ByteDance}}
\affiliation[3]{\mbox{Shanghai Innovation Institute}}

\correspondence{\email{yxcao@fudan.edu.cn}}

\abstract{

}
\begin{document}
\maketitle

\begingroup
\renewcommand{\thefootnote}{}
\footnotetext{* Contributed equally.}
\footnotetext{$\dagger$ Corresponding author.}
\endgroup

\section{Introduction}
\label{sec:introduction}

Language models first made intermediate computation visible through chain-of-thought prompting and self-consistency sampling~\citep{wei2022cot,wang2023selfconsistency}.
Agent benchmarks extend this trace to interleaved thoughts, tool calls, and environment feedback, as in ReAct and Toolformer-style tool use~\citep{yao2023react,schick2023toolformer}.
Yet most reports still collapse each rollout to one scalar: a pass/fail label, reward score, or leaderboard average.
Such scores rank systems, but they do not reveal what process a benchmark rewards.

The challenge is that these process differences are not exposed as aligned units. Agent trajectories are long, branching, and benchmark-specific; their states mix tool choices, observations, file cues, environment feedback, and partial evidence. A single rollout can be inspected manually, but such inspection does not tell us whether another model reached an analogous state, avoided the same failure region, or recovered through a different route. What is missing is a shared coordinate system for comparing many trajectories as movements through the same task landscape.

\begin{wrapfigure}[20]{r}{0.36\linewidth}
  \vspace{-10pt}
  \centering
  \includegraphics[
    width=\linewidth,
  ]{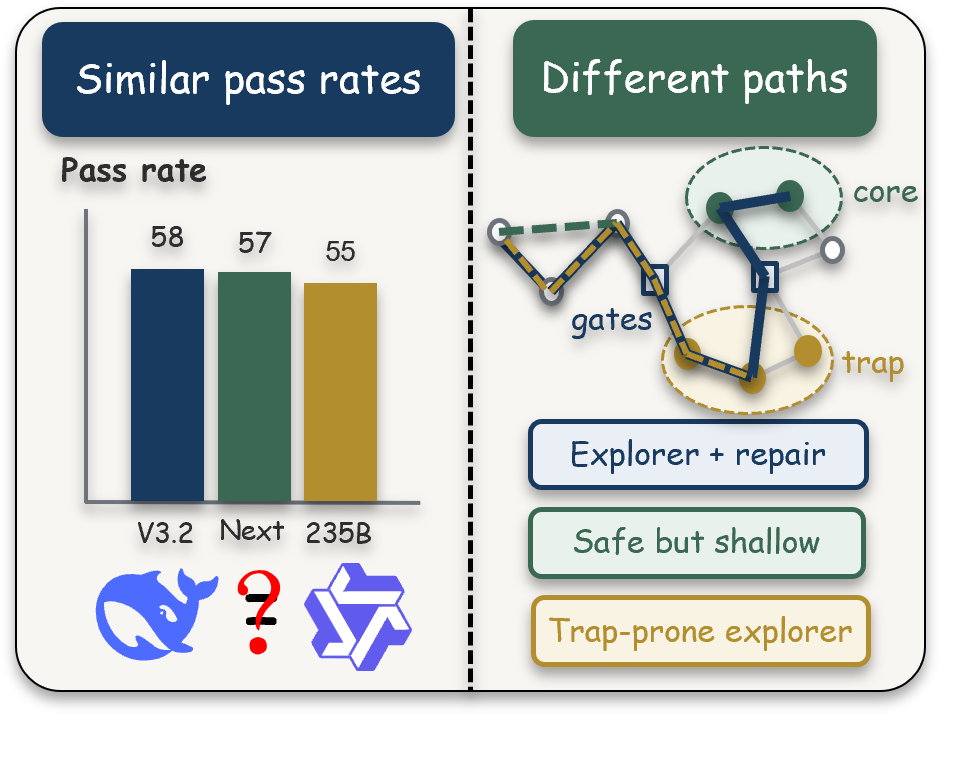}
  \vspace{-8pt}
  \caption{Pass rates can hide process diversity. \textbf{Left}: three models score within 3 points on the same benchmark. \textbf{Right}: on a shared decision landscape built from pooled rollouts, they follow distinct paths: broad exploration with repair, shallow trap avoidance, or high-access navigation with more trap exposure.}
  \label{fig:teaser}
  \vspace{-12pt}
\end{wrapfigure}

We introduce \method{}, a framework that analyzes agent behavior not by inspecting individual trajectories in isolation, but by placing many trajectories from different models onto a shared graph-based decision landscape using observable action-observation signatures. The key hypothesis is that, once trajectories are aligned through observable action–observation states, model behaviors are not arbitrary traces: they form recurring traffic patterns over task-specific regions. 
\method{} then overlays outcome-informed productive cores and trap regions, and summarizes each rollout with three events: reaching a core (\emph{Access}), entering a trap (\emph{Trap exposure}), and returning from a trap to a core (\emph{Repair}).
The name reflects the workflow: traces become graphs, and models are compared by how they move through the resulting landscape.
These roles use terminal outcomes by design, so \method{} is descriptive rather than a blind success predictor: given a common map, what process does each model supply, and what process does each benchmark demand?

A shared landscape is useful not only for retrospective diagnosis, but also for deciding when a live agent trajectory has entered a historically risky region.
If trap regions correspond to repeatable trajectory states, an online detector built from those states should identify moments where a small recovery policy can help.
We therefore instantiate a \method{}-guided trap-aware prefix-fork pipeline on the benchmark with the clearest positive Repair demand.
An agent proceeds until a trap detector fires; matched continuations then resume from the same state while varying a conservative diagnosis note and a temperature bump.
The goal is not to claim a universal recovery mechanism, but to show that graph-derived traps can guide a downstream policy that improves task resolution.

Our contributions can be summarized:
\begin{itemize}
\item \textbf{\method{} for shared decision landscapes.}
We construct model-agnostic task graphs from pooled rollouts, then overlay outcome-informed core and trap regions after graph construction.
\item \textbf{A compact process profile.}
Across a five-split trajectory release spanning 7,329 records, 427 tasks, and five models, Access, Trap exposure, and Repair separate model navigation styles and benchmark demands that aggregate scores collapse.
\item \textbf{A \method{}-guided recovery pipeline.}
Because \swe{} shows the clearest positive Repair demand, we turn graph-derived traps into runtime triggers and test two single-factor recovery policies on all 500 \swe{} Verified instances. The best pooled policy improves resolved rate by $+3.1$ points ($p{=}0.016$) on the per-provider fired subset and by $+3.8$ points ($p{=}0.045$) on common-fired instances, while the active component differs by provider.
\end{itemize}

\section{Related Work}
\label{sec:related_work}

\paragraph{Outcome-centered evaluation.}
Frameworks such as HELM and the LM Evaluation Harness emphasize coverage, transparency, and reproducibility across many tasks and metrics~\citep{liang2023helm,gao2023lmeval}.
Agent benchmarks bring this apparatus to interactive settings, including software repair, tool use, and general agent environments~\citep{jimenez2024swebench,liu2024agentbench,yao2025taubench}.
They are indispensable for tracking progress, but their headline artifacts usually summarize final success.
We reuse the released logs for a complementary purpose: describing the process demands that separate high- and low-outcome trajectories.

\paragraph{Process traces, failure patterns, and intervention.}
Chain-of-thought and self-consistency showed that sampled traces expose alternative reasoning paths before answers are aggregated~\citep{wei2022cot,wang2023selfconsistency}.
ReAct, Toolformer, WebShop, and ALFWorld extended this process view to tool use and environment interaction~\citep{yao2023react,schick2023toolformer,yao2022webshop,shridhar2021alfworld}.
Recent intervention work also treats long-trace failures as actionable: Trap-Aware Adaptive Restart identifies ``thinking traps'' and restarts from earlier reasoning prefixes~\citep{chen2026taar}.
We study an analogous but distinct setting: tool-using software agents, where trap states are observable action--observation signatures and continuations resume from a live benchmark state.

\paragraph{Graphs as inference programs and measurement objects.}
Tree-of-Thoughts and Graph-of-Thoughts use graph structure during inference to guide search~\citep{yao2023tree,besta2024graph}.
Other work uses graphs after generation to cluster chain-of-thought units, analyze trace topology, and model software-agent processes over exploration, localization, patching, validation, and loops~\citep{xiong2025mapping,tan2025shape,liu2026graphectory}.
A related methodological precedent is \sg{}, a post-hoc atlas of same-answer chain-of-thought route isomerism over hidden activation slices~\citep{chen2026slicegraph}.
\method{} adopts this graph-as-measurement perspective, but shifts the object to observable agent states, multi-model task graphs, benchmark-level process demand, and trap-triggered recovery on \swe{}.

\section{\method{}: Building Shared Decision Landscapes}
\label{sec:method}

\begin{figure*}[t]
\centering
\includegraphics[width=\textwidth]{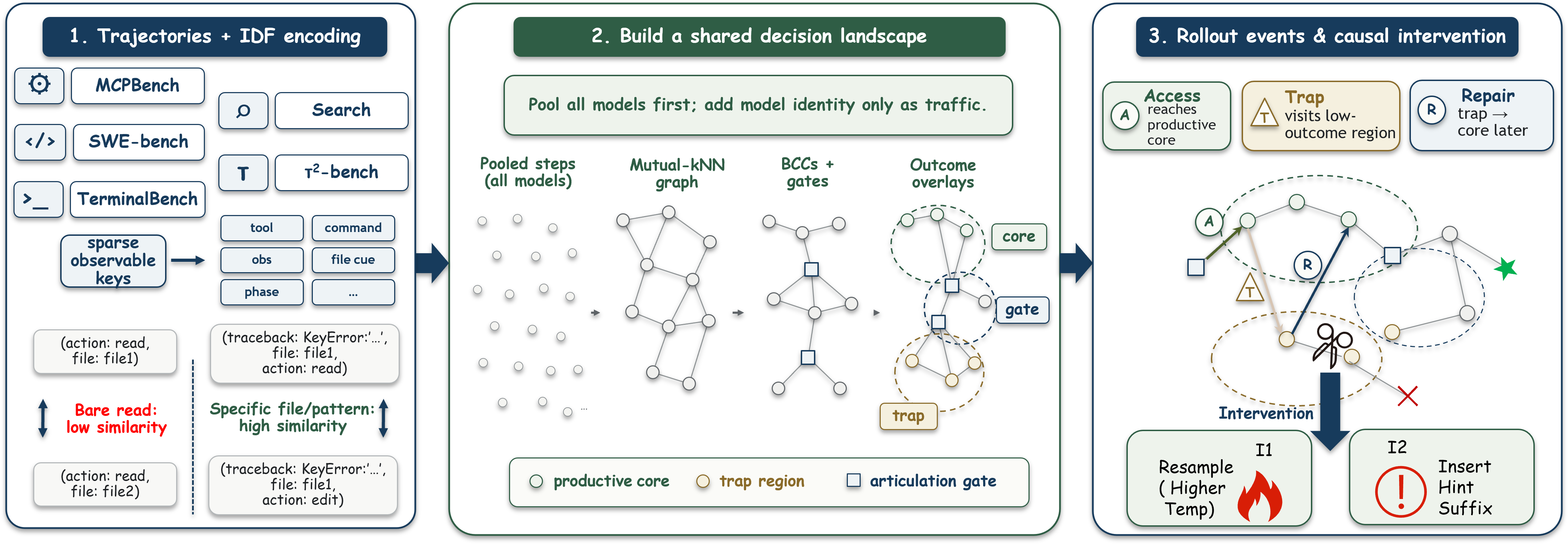}
\caption{Overview of \method{}. \textbf{Stage~1}: action--observation steps from five benchmark splits are encoded into sparse observable keys and compared via IDF-weighted Jaccard similarity. \textbf{Stage~2}: all model rollouts are pooled into a mutual-$k$NN graph, decomposed into BCCs and articulation gates, then annotated with outcome-informed overlays (productive cores, trap regions). \textbf{Stage~3}: three rollout events---Access, Trap, and Repair---are read from each model's path. The same trap overlay later seeds a runtime trigger for a lightweight \swe{} recovery pipeline.}
\label{fig:pipeline}
\end{figure*}

\method{} takes as input a set of agent rollouts and returns a shared decision landscape with three readable rollout events.
For a task $t$, let $\mathcal{R}_t=\bigcup_m\mathcal{R}_{t,m}$ be all rollouts from all available models.
Because the released data contain only a few passes per model per task, \method{} builds one landscape from $\mathcal{R}_t$ and compares models only after that landscape is fixed.
This separation is central to the framework: graph construction defines the terrain, while model and benchmark analyses are overlays on that terrain.

\subsection{Stage 1: Building the Shared Landscape}
\label{sec:shared_graph}

Each action--observation step $i$ is encoded as a sparse set of observable keys $K_i$: tool and action type, command class, observation pattern, file cue, temporal phase, and a pre-specified search-interaction family for \searchb{}.
Pairwise similarity is IDF-weighted Jaccard overlap over key sets,
\begin{equation}
\mathrm{sim}(i,j)=\frac{\sum_{k\in K_i\cap K_j}\idf(k)}{\sum_{k\in K_i\cup K_j}\idf(k)},
\label{eq:sim}
\end{equation}
so rare and more informative keys like a specific traceback pattern at a specific file weigh more than generic ones like a bare ''read'' action.
We convert this to a distance $d(i,j)=1-\mathrm{sim}(i,j)$; among mutual-$k$NN pairs, the graph edge weight is $w_{ij}=\exp(-d(i,j)/\sigma)$ with fixed scale $\sigma$.
For each task we build this mutual-$k$NN graph over all pooled steps, then decompose it into biconnected components (BCCs) and articulation points.
The BCC decomposition is the structural foundation of the landscape: each BCC groups steps that share dense mutual similarity, and the \emph{articulation points} (or \emph{gates}) between BCCs mark strategy or phase transitions---places where a rollout can branch into qualitatively different continuations.
A separate annotation pass confirms that 82 of 100 graph-identified articulation crossings correspond to a human-judged strategy or phase change ($4.6\times$ enrichment over random consecutive steps; see \S\ref{sec:rq1}).
Model identity is never part of the signature, distance, or graph construction.
The exact key alphabet and fixed graph hyperparameters are in Appendix~\ref{app:keys} and Appendix~\ref{tab:hyperparams_full}.

\subsection{Stage 2: Outcome Overlays}
\label{sec:overlays}

On top of the fixed BCC structure, we add two outcome-informed overlays.
Let $G_B=(\mathcal{B}_t,E_B)$ be the block quotient graph, whose nodes are BCC blocks.
For a block $b$, let $\mathcal{R}_b=\{r:\,b\in V_r\}$ be the rollouts that visit it and let $y_r\in[0,1]$ be the terminal outcome (binary for all benchmarks except the max-normalized \mcp{} reward).
We assign each block a Laplace-smoothed reward,
\begin{equation}
\hat\mu_b=\frac{\sum_{r\in\mathcal{R}_b} y_r + \tfrac{1}{2}}{|\mathcal{R}_b|+1},
\qquad
s_b=\hat\mu_b-\bar y_t,
\label{eq:block_reward_main}
\end{equation}
where $\bar y_t$ is the mean outcome over rollouts of task $t$.
To avoid treating each block independently, we diffuse this centered reward over the quotient graph:
\begin{equation}
v^{(\ell+1)}=\alpha P_B^{\top}v^{(\ell)}+(1-\alpha)s,
\label{eq:reward_diffusion_main}
\end{equation}
where $P_B$ is the row-normalized weighted adjacency matrix of $G_B$ and $v^{(0)}=s$.
After fixed-step diffusion and $L_\infty$ normalization, the productive core and trap masks are the high positive and low negative quantiles,
\begin{align}
C_t &= \{b:\, v_b>0 \;\wedge\; v_b\ge Q^+_{.75}\}, \nonumber\\
B_t &= \{b:\, v_b<0 \;\wedge\; v_b\le Q^-_{.25}\},
\label{eq:core_trap_main}
\end{align}
where $Q^+_{.75}$ is the 75th percentile among positive field values and $Q^-_{.25}$ is the 25th percentile among negative field values.
These overlays are descriptive: they locate where high- and low-outcome traffic concentrates, rather than predicting outcomes for unseen runs.
Thus $C_t$ and $B_t$ are not semantic labels like ``good action'' or ``bad action''; a trap block can still contain exploratory traffic (see \S\ref{sec:rq1}).

\subsection{Stage 3: Rollout Events}
\label{sec:profiles}

The final \method{} representation uses only three rollout events, all defined on core/trap overlays.
For a rollout $r$, let $V_r$ be its visited block set and $b_{r,1:T_r}$ its compact block sequence.
\emph{Access} asks whether $V_r$ intersects $C_t$.
\emph{Trap exposure} asks whether $V_r$ intersects $B_t$.
\emph{Repair} asks whether the sequence visits a trap and later reaches a core.
\begin{align}
A_r &= \mathbf{1}[V_r\cap C_t\ne\emptyset], \nonumber\\
E_r &= \mathbf{1}[V_r\cap B_t\ne\emptyset], \nonumber\\
R_r &= \mathbf{1}[\exists s<u:\, b_{r,s}\in B_t,\ b_{r,u}\in C_t]. \label{eq:events_main}
\end{align}
For binary-reward benchmarks, the rollout outcome is the recorded binary success/reward field; for \mcp{}, it is the per-task max-normalized continuous reward.
Appendix~\ref{app:rollout_events} gives full definitions.

\subsection{Supply and Demand Profiles}
\label{sec:supply_demand}

\method{} turns the three events into two descriptive summaries.
A model's \emph{supply} is the task-centered mean of its rollout events,
\begin{equation}
S_{m,a}=\mathbb{E}_t\bigl[a_{m,t}-\mathbb{E}_{m'\in M_t}a_{m',t}\bigr],
\end{equation}
where $a\in\{A,E,R\}$ and $a_{m,t}$ is the average event value for model $m$ on task $t$.
A benchmark's \emph{demand} is the reward-weighted contrast between high- and low-outcome traffic,
\begin{align}
D_{b,a}
&=\frac{\sum_{r\in b} y_r a_r}{\sum_{r\in b}y_r}
 -\frac{\sum_{r\in b}(1-y_r)a_r}{\sum_{r\in b}(1-y_r)}. \label{eq:demand_main}
\end{align}
For binary rewards this is simply the mean event value among resolved rollouts minus the mean event value among failed rollouts.
For \mcp{} it is the continuous analogue using normalized reward.
We use these profiles for interpretation, not as a score-prediction model.

\section{Graph Analysis with \method{}}
\label{sec:experiments}

\subsection{Data and Setup}
\label{sec:data}

\begin{table}[H]
\centering
\scriptsize
\setlength{\tabcolsep}{3pt}
\begin{tabularx}{\linewidth}{lrrX}
\toprule
\textbf{Split} & \textbf{Records} & \textbf{Models} & \textbf{Process focus} \\
\midrule
\mcp{} & 899 & 5 & MCP tool orchestration \\
\searchb{} & 3,270 & 5 & Web search and answer synthesis \\
\swe{} & 747 & 5 & Repository editing and tests \\
\taub{} & 984 & 5 & API use under policies \\
\terminal{} & 1,429 & 5 & Command-line tasks \\
\bottomrule
\end{tabularx}
\caption{Trajectory records in the five retained release splits before graph-validity filtering. Each task is attempted by five models with up to roughly four passes per model, with some missing or filtered attempts.}
\label{tab:data}
\end{table}

We analyze the gated cx-cmu agent trajectory release on Hugging Face~\citep{cxcmu2026agenttrajectories}.
The analyzed splits include tool orchestration, web-search answer synthesis, software repair, API-policy interaction, and command-line tasks from \mcp{}~\citep{wang2025mcpbench}, the \searchb{} split of the cx-cmu trajectory release~\citep{cxcmu2026agenttrajectories}, \swe{}~\citep{jimenez2024swebench}, \taub{}~\citep{barres2025tau2}, and \terminal{}~\citep{merrill2026terminalbench}.
The public dataset card describes several benchmark splits and multiple passes from the same five models on the same tasks; it records \searchb{} as a split label with web-search trajectories for answering complex questions, including \texttt{browsecomp}, \texttt{webvoyager}, and \texttt{mind2web} domains~\citep{cxcmu2026agenttrajectories}. We therefore use \searchb{} as the release split name rather than as a standalone benchmark title, while noting that the release tool inventory is reconstructed from the General-AgentBench source tree~\citep{cxcmu2026agenttrajectories,li2026generalagentbench}.
We use the five splits with rich tool or environment interaction and exclude the short math split, leaving a retained release subset of 7,329 trajectory records on 427 tasks before graph-validity filtering.
The analyzed trajectories are English-language task interactions with code, tool, or API observations.
The models are DeepSeek-R1, DeepSeek-V3.2, Gemini-2.5-Flash, Qwen3-235B, and Qwen3-Next.

For the four binary-reward splits, $y_r$ is the recorded binary success/reward flag: for example, \swe{} uses resolved-style issue success, whereas \searchb{} uses the binary \texttt{score} recorded in the release.
For \mcp{}, which reports a continuous multi-criteria score, we use per-task max-normalized reward.
This avoids introducing a binary threshold for a split that is near-saturated under a positive-reward rule.
For \searchb{}, we add pre-specified search-interaction keys for URL domain, query novelty, and cumulative source diversity; exact extraction rules are in Appendix~\ref{app:keys}.

\begin{figure*}[t]
\centering
\begin{subfigure}[t]{0.32\textwidth}
\includegraphics[width=\textwidth]{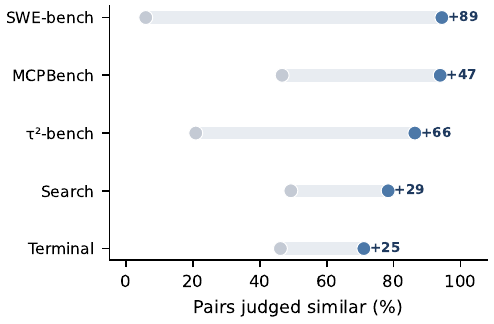}
\end{subfigure}\hfill
\begin{subfigure}[t]{0.32\textwidth}
\includegraphics[width=\textwidth]{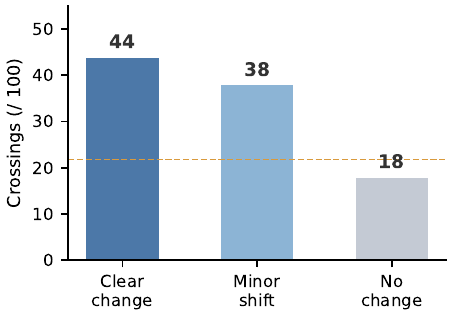}
\end{subfigure}\hfill
\begin{subfigure}[t]{0.32\textwidth}
\includegraphics[width=\textwidth]{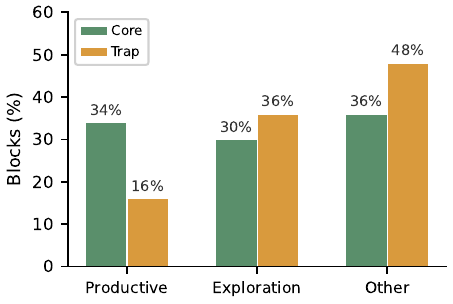}
\end{subfigure}
\caption{Human annotation checks for graph-derived states. \textbf{(a)}~Within-BCC step pairs are judged similar far more often than across-BCC pairs on every benchmark (overall $84.8\%$ vs. $40.2\%$ after dropping unsure labels; Fisher OR$\,{=}\,8.30$). \textbf{(b)}~Of 100 graph-identified articulation crossings, 82 show a strategy or phase change ($4.6\times$ the random baseline, dashed line). \textbf{(c)}~Core blocks are annotated as productive progress $2.1\times$ more often than trap blocks ($34\%$ vs. $16\%$); $36\%$ of trap blocks are exploration rather than dead ends.}
\label{fig:rq1_validity}
\end{figure*}

\subsection{Do \method{} Landscapes Expose Interpretable Agent States?}
\label{sec:rq1}

We first check whether a \method{} landscape is a usable measurement scaffold.
The evidence is deliberately modest: it asks whether the graph units are interpretable enough to support analysis, not whether every block has a perfect semantic label.

\paragraph{Block semantics.}
Two annotators scored 400 blinded BCC pairs, split evenly between within-BCC and across-BCC pairs.
After dropping unsure labels, within-BCC pairs were marked similar 84.8\% of the time (162/191), compared with 40.2\% for across-BCC pairs (72/179; Fisher exact OR=8.30, $p=1.4\times 10^{-19}$).
Per-benchmark gaps range from +89 percentage points on \swe{} to +25 percentage points on \terminal{}; \searchb{} has a +29 point gap under the pre-enrichment signature.
This supports reading BCCs as coarse shared agent states (Figure~\ref{fig:rq1_validity}a).

\paragraph{Articulations and regions.}
A second annotation pass scored 100 graph-identified articulation crossings.
Eighty-two show at least a minor strategy or phase change, a 4.6$\times$ enrichment over random consecutive steps (Figure~\ref{fig:rq1_validity}b).
We use this as a graph-construction sanity check, not as a main supply--demand axis.
A third pass scored 50 core blocks and 50 trap blocks.
Core blocks are 2.1$\times$ more likely than trap blocks to be judged productive progress (34\% vs. 16\%, Fisher $p=0.032$), and region-consistent extreme labels---productive core blocks and looping/dead-end trap blocks---outnumber inconsistent extremes 35 to 11.
However, 36\% of trap blocks are judged exploration, so a trap is best understood as a low-outcome region, not necessarily useless behavior (Figure~\ref{fig:rq1_validity}c).

\paragraph{Why a shared graph?}
Pooled graphs produce an average of 16.1 non-trivial blocks across 408 valid tasks, enough to compare several models as traffic on the same terrain.
Per-model graphs are much smaller because each model contributes only a few rollouts per task.
A representation ablation further shows that the full multi-channel signature has the broadest valid-task coverage; the ablation is kept in Appendix~\ref{app:signature_ablation} because it is a sanity check rather than the semantic core of the paper.

\subsection{What Process Profiles Do Models Exhibit?}
\label{sec:rq2}

Once the landscape is fixed, model identity is traffic over the same terrain.
Figure~\ref{fig:supply_demand}a shows model supply as a heatmap; exact values with bootstrap CIs are in Appendix Table~\ref{tab:model_profiles}.
Positive values mean the model exhibits that event more often than the average model on the same tasks; for Trap exposure, lower values mean more avoidance.

The profiles give model descriptions rather than a new ranking.
\textbf{DeepSeek-V3.2} is the clearest high-access, high-repair model: its Access and Repair CIs are both positive, while Trap is nearly neutral.
\textbf{Qwen3-Next} is low-access and low-repair, with lower Trap supply; this looks less like robust recovery and more like conservative, shallow navigation.
\textbf{Qwen3-235B} shows mildly elevated Access and marginally elevated Trap exposure, suggesting a less conservative exploration style, but its Repair interval overlaps zero.
\textbf{DeepSeek-R1} is near-balanced, and \textbf{Gemini-2.5-Flash} has lower Access and Repair supply in this corpus.

\begin{figure*}[t]
\centering
\begin{subfigure}[t]{0.48\textwidth}
\includegraphics[width=\textwidth]{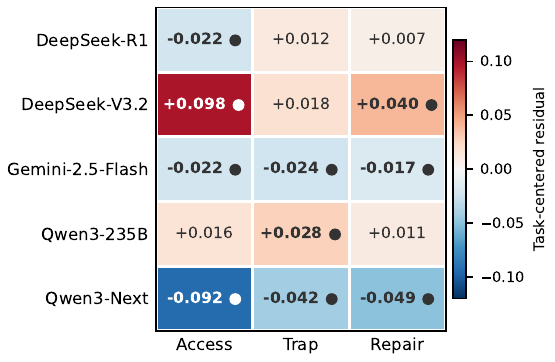}
\end{subfigure}\hfill
\begin{subfigure}[t]{0.48\textwidth}
\includegraphics[width=\textwidth]{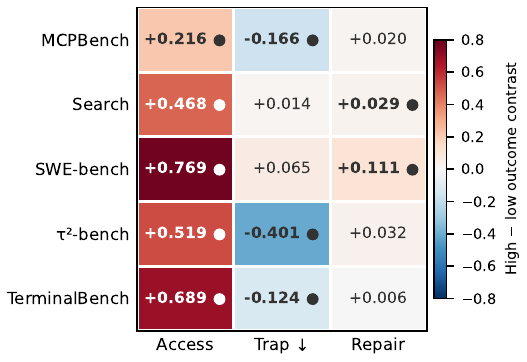}
\end{subfigure}
\caption{Supply--demand profile heatmaps. \textbf{(a)}~Model supply: task-centered event residuals; DeepSeek-V3.2 is distinctly high-Access/high-Repair, while Qwen3-Next is low on both. \textbf{(b)}~Benchmark demand: resolved-minus-failed contrasts (or the reward-weighted analogue on \mcp{}); Access is universally positive, \taub{} is strongly Trap-averse, \mcp{}/\terminal{} are also negative on Trap, and \swe{} has by far the strongest positive Repair demand. Filled markers~({\large$\bullet$}) denote 95\% bootstrap CIs excluding zero.}
\label{fig:supply_demand}
\end{figure*}

\begin{figure*}[t]
\centering
\begin{subfigure}[t]{0.32\textwidth}
\includegraphics[width=\textwidth]{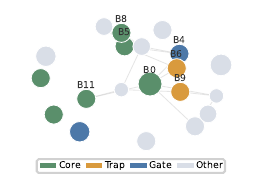}
\end{subfigure}\hfill
\begin{subfigure}[t]{0.32\textwidth}
\includegraphics[width=\textwidth]{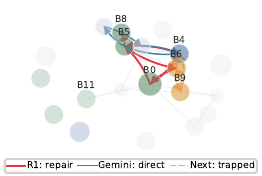}
\end{subfigure}\hfill
\begin{subfigure}[t]{0.32\textwidth}
\includegraphics[width=\textwidth]{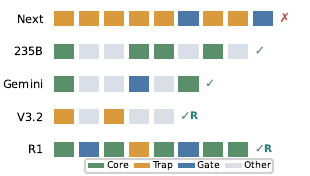}
\end{subfigure}
\caption{A shared landscape on \swe{} task \texttt{django-9296}. \textbf{(a)}~Block quotient graph: nodes are BCC blocks ({\color[HTML]{5A8F6B}green}=core, {\color[HTML]{D99A3D}amber}=trap, blue=articulation landmark, gray=other). \textbf{(b)}~Three trajectories as arrows: {\color[HTML]{E63946}DeepSeek-R1} repairs from B6 back to core; {\color[HTML]{457B9D}Gemini} navigates directly; {\color[HTML]{F4A261}Qwen3-Next} (dashed) cycles in B9$\leftrightarrow$B6 without escaping. \textbf{(c)}~Block ribbons per model; \checkmark/\texttimes\ marks resolution, \textbf{R} marks Repair.}
\label{fig:case_study}
\end{figure*}

\subsection{What Process Demands Do Benchmarks Impose?}
\label{sec:rq3}

We next ask which events separate high-outcome from low-outcome traffic inside each benchmark.
Figure~\ref{fig:supply_demand}b shows the demand heatmap; exact values with bootstrap CIs are in Appendix Table~\ref{tab:demand_vectors}.
On binary splits this is resolved minus failed traffic; on \mcp{} it is the continuous reward-weighted analogue.

Access is broadly shared: every split has a positive Access demand with CIs excluding zero.
The more diagnostic differences are Trap and Repair.

\paragraph{Trap-averse benchmarks.}
\taub{} has the most negative Trap demand ($-0.401$), while \mcp{} ($-0.166$) and \terminal{} ($-0.124$) are also trap-averse with intervals below zero.
High-outcome rollouts in these splits avoid low-outcome regions, which matches settings where state updates, protocol mistakes, or invalid tool choices can be difficult to undo.
\mcp{} also has a clearly positive Access demand ($+0.216$), suggesting that successful tool-orchestration trajectories must both reach productive regions and avoid drifting into low-outcome states.
This explains why lower-Trap supply can be useful even when a model is otherwise low-access.

\paragraph{Repair-heavy software tasks.}
\swe{} has the strongest Repair demand ($+0.111$) and the strongest Access demand ($+0.769$), both with positive intervals.
Successful repository-editing trajectories must both reach productive code regions and recover after failed tests, wrong files, or bad edits.
This makes \swe{} the natural setting for the \method{}-guided recovery pipeline in \S\ref{sec:trap_intervention}: the intended lever is not trap avoidance, but recovery after a trap-like state.
By contrast, the negative Trap demand on \taub{} and \mcp{} suggests a different design target, namely preventing risky state updates before they occur, so we do not use a repair-style diagnosis policy as the pipeline test for those splits.

\paragraph{Search as access-and-repair.}
\searchb{} is not trap-averse in this signature, but it has positive Access demand ($+0.468$) and a smaller positive Repair demand ($+0.029$).
Under the evidence-quality keys, high-outcome search rollouts are distinguished mainly by reaching productive evidence states, with occasional recovery from unproductive query paths.

Together, the supply and demand profiles make the central point: benchmarks are not simply harder or easier versions of the same task.
They reward different process events, and models align with those demands in different ways.
Figure~\ref{fig:supply_demand} visualises both matrices as heatmaps; filled markers denote bootstrap CIs excluding zero.
Figure~\ref{fig:case_study} concretizes this on a single \swe{} task, showing how three models navigate the same shared landscape with visibly different strategies.

\section{\method{}-Guided Trap-Aware Recovery}
\label{sec:trap_intervention}

The \method{} analysis asks where high- and low-outcome traffic concentrates on a shared task map.
We next test whether that map can guide a small recovery pipeline: act only when a live rollout reaches a state resembling historical traps, and use only information already visible in the agent's context.
Because \S\ref{sec:rq3} shows that \swe{} most strongly demands recovery after trap exposure, we instantiate trap-aware recovery on \swe{} and measure official resolved rate after trap-triggered continuations.
The goal is not a universal repair algorithm, but a test of whether graph-derived traps identify states where a lightweight, non-oracular continuation change helps.

\begin{table*}[!htbp]
\centering
\small
\setlength{\tabcolsep}{5pt}
\renewcommand{\arraystretch}{1.15}
\begin{tabular}{lc|c|cc}
\toprule
\textbf{Provider} & $n_{\text{fired}}$ & \textbf{Baseline} & \textbf{Hot} & \textbf{Note} \\
 & & {\scriptsize($T{=}0.6$, no note)} & {\scriptsize($T{=}0.9$, no note)} & {\scriptsize($T{=}0.6$, note)} \\
\midrule
\multicolumn{5}{c}{\emph{Per-provider fired subset (each provider's own complete set)}} \\
\midrule
Qwen3.6-35B-A3B & 214 & 29.4 & 30.8 {\scriptsize\textcolor{gaincolor}{(+1.4)}} & \cellcolor{bestcolor}\textbf{34.1} {\scriptsize\textcolor{gaincolor}{(+4.7)}} \\
GLM-5.1 & 196 & 48.0 & \cellcolor{bestcolor}\textbf{52.6} {\scriptsize\textcolor{gaincolor}{(+4.6)}} & 49.5 {\scriptsize\textcolor{gaincolor}{(+1.5)}} \\
DeepSeek-V4-Pro & 206 & 44.7 & \cellcolor{bestcolor}\textbf{48.1} {\scriptsize\textcolor{gaincolor}{(+3.4)}} & 45.6 {\scriptsize\textcolor{gaincolor}{(+1.0)}} \\
\midrule
\rowcolor{highlight}
Pooled (3-prov, $n{=}616$ cells) & 616 & 40.4 & \textbf{43.5} {\scriptsize\textcolor{gaincolor}{(+3.1)}} & 42.9 {\scriptsize\textcolor{gaincolor}{(+2.4)}} \\
\midrule
\multicolumn{5}{c}{\emph{Common-fired subset (same 96 instances, all three providers)}} \\
\midrule
Qwen3.6-35B-A3B & 96 & 31.2 & 31.2 \phantom{{\scriptsize{(+0.0)}}} & \cellcolor{bestcolor}\textbf{34.4} {\scriptsize\textcolor{gaincolor}{(+3.1)}} \\
GLM-5.1 & 96 & 50.0 & \cellcolor{bestcolor}\textbf{56.2} {\scriptsize\textcolor{gaincolor}{(+6.2)}} & 52.1 {\scriptsize\textcolor{gaincolor}{(+2.1)}} \\
DeepSeek-V4-Pro & 96 & 41.7 & \cellcolor{bestcolor}\textbf{46.9} {\scriptsize\textcolor{gaincolor}{(+5.2)}} & 44.8 {\scriptsize\textcolor{gaincolor}{(+3.1)}} \\
\midrule
\rowcolor{highlight}
Pooled (3-prov, $n{=}288$ cells) & 288 & 41.0 & \textbf{44.8} {\scriptsize\textcolor{gaincolor}{(+3.8)}} & 43.8 {\scriptsize\textcolor{gaincolor}{(+2.8)}} \\
\bottomrule
\end{tabular}
\caption{\method{}-guided trap-aware recovery on the full 500-instance \swe{} Verified pool. Cells report official resolved rate (\%); parentheses show paired gain over Baseline. Gains are computed from unrounded paired outcomes, so they may differ by 0.1 pp from subtracting displayed rounded rates. Best non-Baseline cells are bold-shaded. Appendix~\ref{app:trap_intervention_more} reports uncertainty; Appendix~\ref{app:trap_case_studies} gives cases.}
\label{tab:swe_4arm_intervention}
\end{table*}

\subsection{From Trap Regions to Runtime Triggers}
\label{sec:intervention_design}

\method{} converts the trap overlay into a runtime detector by storing canonicalized key sets from trap-side states, such as \texttt{ACTION:edit}, \texttt{OBS:error}, \texttt{TOOL:<name>}, and local file-path cues.
Generic phase and non-tool keys are stripped so the detector matches recurring failure contexts rather than full traces.
A second library comes from high-committor non-trap states.
At a live step $s$, the detector computes the candidate key set $q_s$ and scores
\begin{align}
\mathrm{sim}_{\mathrm{trap}}(s) &= \max_{u\in\mathcal{T}} J_{\idf}(q_s,u), \nonumber\\
\mathrm{sim}_{\mathrm{core}}(s) &= \max_{u\in\mathcal{N}} J_{\idf}(q_s,u),
\end{align}
where $\mathcal{T}$ and $\mathcal{N}$ are the trap and non-trap libraries.
A trigger fires when trap similarity clears a fixed threshold and exceeds core similarity by a margin (Appendix Table~\ref{tab:detector_hparams}).
For \swe{}, the current step must also carry a local file-path cue, and the candidate action must be an edit or submit action.
These local checks keep the trigger tied to observable graph signatures while reducing fires on generic exploration.

\subsection{Prefix-Fork Recovery Policy}
\label{sec:prefix_fork}

When the detector fires, we snapshot the docker workspace and conversation prefix, then continue from that identical state under three policies.
The no-intervention \emph{Baseline} continues at $T{=}0.6$ with no added note.
\emph{Hot} raises temperature to $T{=}0.9$ with top-$p{=}0.95$.
\emph{Note} keeps $T{=}0.6$ and appends a conservative diagnosis note drawn only from the agent's recent log: the last action, observation signal, implicated file or target keys, detector confidence, and a family-level diagnosis.
The note asks the agent to re-read the failing evidence, localize narrowly, make one minimal change, and revise or discard the edit if the evidence does not support it.
No arm receives a gold patch, verified-test list, or oracle information.

This prefix-fork design compares the same trap state with no pipeline action, extra sampling diversity, or evidence-grounded localization.
We report official \swe{} resolved rate in Table~\ref{tab:swe_4arm_intervention}; Appendix~\ref{app:trap_intervention_more} provides paired uncertainty estimates.

\subsection{Downstream Accuracy Improves at Trap-Triggered States}
\label{sec:swe_intervention}

The experiment uses the full 500-instance \swe{} Verified pool at seed $11$, spanning 12 repositories and three continuation providers not used in the observational graph analysis: Qwen3.6-35B-A3B, GLM-5.1, and DeepSeek-V4-Pro.
Table~\ref{tab:swe_4arm_intervention} reports each provider's complete fired subset and a common-fired subset of 96 instances where all providers fire and complete all three arms.

\method{}-triggered recovery improves downstream resolution on fired states.
In the per-provider fired subset, pooled Hot raises resolved rate from $40.4\%$ to $43.5\%$ ($+3.1$ pp, $p{=}0.016$), and pooled Note reaches $42.9\%$ ($+2.4$ pp, $p{=}0.064$).
The common-fired subset shows the same pattern: Hot increases pooled resolved rate from $41.0\%$ to $44.8\%$ ($+3.8$ pp, $p{=}0.045$).
Thus a trap region discovered from shared offline trajectories can become a runtime trigger that improves a downstream metric.

The active recovery route is provider-specific: Qwen3.6-35B-A3B benefits most from the diagnosis note ($+4.7$ pp, $p{=}0.036$), while GLM-5.1 ($+4.6$ pp, $p{=}0.037$) and DeepSeek-V4-Pro ($+3.4$ pp) benefit most from the temperature bump.

Appendix~\ref{app:trap_intervention_more} reports the full component statistics and the conservative diagnosis template; Appendix~\ref{app:trap_case_studies} gives case studies showing how the same trigger state can lead to different provider-specific recoveries.

\section{Discussion and Conclusion}
\label{sec:discussion}

\method{} is neither another scalar score nor a blind success predictor.
Its outcome-informed core and trap regions are descriptive: they show where successful and failed traffic concentrates on a shared map, turning each rollout into a process reading over Access, Trap exposure, and Repair.
This exposes differences hidden by aggregate scores.
DeepSeek-V3.2 is the clearest high-access, high-repair model, Qwen3-Next is conservative and shallow, and other models lie between them (\S\ref{sec:rq2}).
Benchmarks also impose different process demands: \taub{} is strongly trap-averse, \mcp{} and \terminal{} are negative on Trap, \swe{} adds Repair demand, and \searchb{} mainly rewards productive evidence access (\S\ref{sec:rq3}).
Thus, benchmark difficulty is better understood as a mixture of exploration, avoidance, and recovery demands rather than a hidden factor.

The recovery experiment shows that this map can guide a small downstream intervention without becoming a full controller.
Graph-derived traps trigger prefix-fork continuations only when a live \swe{} rollout enters a historical failure region.
On fired states from the 500-instance \swe{} Verified pool, the best pooled single-factor policy raises resolved rate by $+3.1$ points on the per-provider fired subset and by $+3.8$ points on common-fired instances .

\section{Limitations}
\label{sec:limitations}

The \method{} landscape roles are outcome-informed, so the framework is an analytical instrument for explaining where high- and low-outcome traffic concentrates, not a blind success predictor.
The source data contains roughly four passes per model per task across five observational models; shared graphs mitigate sparsity for landscape construction but not for cell-level inference.
For \mcp{}, a per-task max-normalized continuous reward replaces binarization but treats the per-task maximum as the local ceiling, which it may not be.
The symbolic signature captures tool, command, observation, file, and phase cues, but may miss fine-grained natural-language semantics, especially for \searchb{}.
The \method{}-guided recovery pipeline applies only to a structurally selected \swe{} fired subset and three continuation providers, so its gains should be read as local downstream improvements rather than as a universal agent-repair result.

\clearpage
\bibliographystyle{plainnat}
\bibliography{references}

\appendix
\section{Responsible Research Statement}
\label{app:responsible_research}
\paragraph{Potential risks.}
\method{} is an analysis and evaluation framework, not a deployed agent-control system.
The main risks are over-interpreting outcome-informed graph roles as blind predictors of success, overfitting recovery policies to benchmark-specific traps, or using automated software-editing agents without human review.
We mitigate these risks by presenting the recovery pipeline as a local benchmark intervention on fired \swe{} states, using no gold patches or verified-test lists in the recovery note, reporting provider-specific behavior, and avoiding any deployment recommendation.

\paragraph{Artifacts, licenses, and intended use.}
We use access-controlled or provider-served research artifacts: the gated cx-cmu agent trajectory release on Hugging Face, the underlying benchmark suites and evaluation harnesses, and the three continuation models cited in \S\ref{sec:data} and \S\ref{sec:trap_intervention}.
The public card for the cx-cmu release gates file access behind contact sharing, and the public README does not list a separate dataset license or a detailed citation block.
We therefore cite the dataset card directly, treat the trajectories as access-controlled research material, and use them only for the analysis reported in this paper.
For the recovery study, GLM-5.1 and DeepSeek-V4-Pro are accessed through provider APIs under provider terms, while Qwen3.6-35B-A3B is run locally from Qwen-family weights whose public model card lists the Apache-2.0 license.
We do not redistribute raw trajectories, benchmark instances, source repositories, model weights, or API outputs.
Our intended use is research evaluation and process analysis of agent benchmarks, and any future sharing of derived artifacts would need to respect the access conditions of the source artifacts.

\paragraph{Privacy, content, and data handling.}
We do not collect new personal data, infer demographic attributes, or conduct user profiling.
The source artifacts consist of benchmark tasks, model-generated traces, tool outputs, repository paths, issue statements, and synthetic or benchmark-specific environment records.
We inspected the schema and representative examples for fields that directly name or identify people, and the analysis converts textual fields into canonicalized symbolic keys such as tool type, command class, observation pattern, file cue, and phase cue.
The paper reports aggregate statistics and illustrative benchmark snippets rather than raw private user conversations.

\paragraph{Computation and reproducibility.}
We do not train or fine-tune models.
Graph construction, annotation analysis, bootstrapping, and table generation use custom scripts over fixed symbolic signatures and fixed hyperparameters reported in \S\ref{sec:method} and Appendix~\ref{tab:hyperparams_full}.
The recovery experiment uses two API-served continuation providers (GLM-5.1 and DeepSeek-V4-Pro) and one locally hosted continuation model (Qwen3.6-35B-A3B).
Provider-side infrastructure for the API models is not observable to us, so we report model/provider names, data sizes, fired subsets, temperatures, top-$p$ values, confidence intervals, and permutation tests rather than provider GPU hours.
The locally hosted Qwen3.6-35B-A3B runs on a single NVIDIA H20 GPU, and the Qwen intervention sweep consumed approximately ten GPU-hours in total.

\paragraph{Human annotation.}
The graph-interpretability sanity checks use two internal annotators rather than crowdworkers or recruited human subjects.
The annotators were given task-level trajectory fragments and asked to judge within- vs. across-block similarity, articulation phase changes, and whether core/trap blocks correspond to productive progress or low-outcome exploration.
No external participants were recruited or paid, no annotator personal data were collected, and the annotation is used only as a sanity check for graph interpretability.
The full instruction text is in Appendix~\ref{app:annotation_guidelines}.

\paragraph{AI assistance.}
AI assistants were used for language polishing, LaTeX editing, patch preparation, and submission-checklist drafting.
The authors supplied the research ideas, experimental design, data analysis, and claims, and manually verified all citations, numerical results, and final text.
No AI assistant is treated as an author.

\section{Additional \method{} Details}
\label{app:details}

\subsection{Hyperparameters}
All graph and diffusion hyperparameters are fixed across benchmarks.

\begin{table}[h]
\centering
\small
\setlength{\tabcolsep}{5pt}
\begin{tabular}{ll}
\toprule
\textbf{Parameter} & \textbf{Value} \\
\midrule
Mutual-$k$NN $k$ & 6 \\
Edge-weight RBF scale $\sigma$ & 0.35 \\
Diffusion teleport $\alpha$ & 0.65 \\
Diffusion steps & 24 \\
Core quantile (top, $C_t$) & $0.75$ \\
Trap quantile (bottom of negative, $B_t$) & $0.25$ \\
Bootstrap resamples & $2{,}000$ \\
\bottomrule
\end{tabular}
\caption{Fixed hyperparameters used by \method{}.}
\label{tab:hyperparams_full}
\end{table}

\subsection{Rollout-Event Definitions}
\label{app:rollout_events}

The three rollout events in \S\ref{sec:profiles} are defined on the role overlays $(C_t,B_t)$.
For a rollout $r$ with compact primary-block sequence $b_{r,1:T_r}$ and visit set $V_r=\{b_{r,s}\}$:

\begin{align}
A_r &= \mathbf{1}[V_r\cap C_t\neq\emptyset], \label{eq:Ar}\\
E_r &= \mathbf{1}[V_r\cap B_t\neq\emptyset], \label{eq:Er}\\
R_r &= \mathbf{1}[\exists s<u:\, b_{r,s}\in B_t,\ b_{r,u}\in C_t]. \label{eq:Rr}
\end{align}

For binary benchmarks $y_r\in\{0,1\}$; for \mcp{} we use the per-task max-normalized continuous reward $y_r=\text{reward}_r/\max_{r'\in\mathcal{R}_t}\text{reward}_{r'}\in[0,1]$.

The model supply $S_{m,a}$ is the average over tasks of the task-centered cell residual:
\begin{align}
a_{m,t} &= \mathbb{E}_{r\in\mathcal{R}_{t,m}}[a_r], \nonumber\\
\tilde a_{m,t} &= a_{m,t}-\frac{1}{|M_t|}\sum_{m'\in M_t}a_{m',t}, \nonumber\\
S_{m,a} &= \mathbb{E}_t[\tilde a_{m,t}],
\end{align}
where $a\in\{A,E,R\}$.

The benchmark demand $D_{b,a}$ is the reward-weighted contrast over rollouts:
\begin{align}
D_{b,a}
&= \frac{\sum_{r\in b} w_r a_r}{\sum_{r\in b} w_r}
 - \frac{\sum_{r\in b} (1-w_r)a_r}{\sum_{r\in b} (1-w_r)}, \nonumber\\
w_r &= y_r.
\label{eq:Dba_app}
\end{align}
For binary $y_r$, $w_r$ is $0/1$ so Eq.~\ref{eq:Dba_app} reduces exactly to the mean-over-resolved minus mean-over-failed contrast.

\subsection{Axis Sign Sanity, Rollout Events}
\label{app:axis_sign_v2}

The three events are descriptive coordinates, not predictors, but they should still align with success in the expected direction across the corpus: high-outcome traffic should have larger $A$ and $R$ and smaller $E$ than low-outcome traffic.
Pooling all rollouts on the four binary splits and computing the unweighted contrast $\E_{y=1}[a_r] - \E_{y=0}[a_r]$ with $2{,}000$-bootstrap $95\%$ CIs, all three events have CIs that exclude zero in the expected direction (Table~\ref{tab:axis_sign_v2}).
Including \mcp{} under the continuous treatment gives the same qualitative pattern: positive Access, negative Trap, and small Repair.

\begin{table}[h]
\centering
\scriptsize
\setlength{\tabcolsep}{3pt}
\begin{tabular}{lcc}
\toprule
\textbf{Event} & \textbf{$\E_{y=1}-\E_{y=0}$} & \textbf{$95\%$ CI} \\
\midrule
Access ($A_r$)             & $+0.572$ & $[+0.544,\,+0.600]$ \\
Trap exposure ($E_r$)      & $-0.064$ & $[-0.091,\,-0.036]$ \\
Repair ($R_r$)             & $+0.040$ & $[+0.023,\,+0.057]$ \\
\midrule
All three sign-consistent  & \multicolumn{2}{c}{Yes} \\
\bottomrule
\end{tabular}
\caption{Rollout-event sign sanity on the four binary benchmarks (4,686 graph-valid binary rollout-event rows; 2,000-bootstrap $95\%$ CIs). All three main events shift in the expected direction at $p<0.05$.}
\label{tab:axis_sign_v2}
\end{table}

\subsection{Signature Ablation}
\label{app:signature_ablation}

\begin{figure}[h]
\centering
\includegraphics[width=\columnwidth]{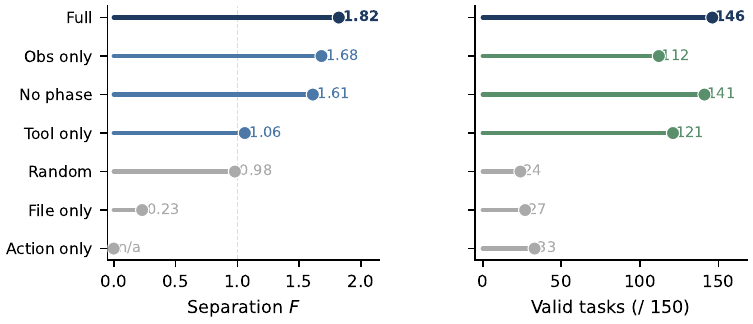}
\caption{Signature ablation on $150$ tasks. Full signatures give the strongest model separation ($F=1.82$) and widest valid-task coverage ($146/150$); random, file-only, and action-only collapse coverage, and action-only yields no stable separation estimate.}
\label{fig:ablation}
\end{figure}

We rebuild the pipeline on $150$ tasks under seven representation conditions.
The full signature has the largest model-separation statistic among non-degenerate representations and the broadest valid-task coverage; observation-only and no-phase variants remain useful, tool-only lowers separation while preserving broad coverage, and file-only, action-only, and random reassignment collapse coverage, with action-only yielding no stable separation estimate (Figure~\ref{fig:ablation}).

\subsection{Score Matrix and Reward Semantics}
\label{app:resolve}

The four binary-reward splits use the resolved/failed flag of each rollout.
For \mcp{}, the dataset provides a continuous multi-criteria reward; the main text uses the per-task max-normalized continuous reward $y_r=\text{reward}_r/\max_{r'\in\mathcal{R}_t}\text{reward}_{r'}$ throughout the reward field and benchmark-demand calculations.
A simple positive-reward threshold is near-saturated ($\sim\!97\%$ resolved across models), so we use the normalized continuous treatment instead; no threshold sweep is reported as a main result because the analysis avoids binarization altogether.

\subsection{Annotation Instructions}
\label{app:annotation_guidelines}

The graph-interpretability sanity checks used two internal annotators.
The annotators were shown de-identified task id, benchmark name, two or three consecutive trajectory fragments, and the graph-derived relation being checked; model identity and terminal outcome were hidden.
The instruction text was:

\begin{quote}
Read only the displayed action, observation, tool, file, and phase cues.
For a pair of steps, label them \emph{similar} if they represent the same local subtask, evidence-gathering mode, tool-use mode, or edit/debugging phase, even if the surface text differs; otherwise label them \emph{different}.
For an articulation crossing, label whether the second step marks a clear strategy or phase change relative to the first.
For a block, label the fragment as productive progress, exploration without convergence, repeated or stuck behavior, or unclear.
Do not infer model identity, do not use terminal success, and mark unclear when the displayed evidence is insufficient.
This annotation is for aggregate method validation only and should not identify any person or judge any human participant.
\end{quote}

No screenshots or additional risk disclaimers were used because the task was an internal, non-interactive annotation of already released agent traces.
The directional sanity-check $p$-values reported in \S\ref{sec:rq1} use one-sided alternatives matched to the stated hypotheses: within-BCC similarity $>$ across-BCC similarity, and core productive-progress rate $>$ trap productive-progress rate.

\subsection{Illustrative Annotation Examples}
\label{app:annotation_cases}

We show representative trajectory fragments from each of the three annotation tasks in \S\ref{sec:rq1}.
Tool arguments and observations are lightly trimmed for space.

\paragraph{BCC pair: within-block (similar).}
\swe{}, task \texttt{sympy-16886}.
Two steps from the same BCC block; the annotator labelled them \emph{similar}.

\smallskip\noindent\small
\begin{tabularx}{\linewidth}{@{}l@{\;\;}X@{}}
\textbf{A} & \texttt{view} \texttt{crypto/crypto.py:1500--1550} $\to$ Morse-code definitions \\[2pt]
\textbf{B} & \texttt{view} \texttt{crypto/crypto.py:500--600} $\to$ Vigen\`ere helpers \\
\end{tabularx}
\normalsize

\smallskip\noindent
Both steps view different sections of the same file, reflecting a single ``inspect source'' strategy.
The graph merges them because their key sets share the high-IDF key \texttt{FILE\_PATH:crypto/crypto.py}.

\paragraph{BCC pair: across-block (different).}
\swe{}, task \texttt{pytest-5773}.
Two steps from different BCC blocks; the annotator labelled them \emph{different}.

\smallskip\noindent\small
\begin{tabularx}{\linewidth}{@{}l@{\;\;}X@{}}
\textbf{A} & \texttt{bash}: \texttt{git show de6f.. | grep "python\_files"} $\to$ commit diff \\[2pt]
\textbf{B} & \texttt{str\_replace} on \texttt{\_pytest/python.py} $\to$ ``replaced'' \\
\end{tabularx}
\normalsize

\smallskip\noindent
Step~A inspects git history; Step~B edits source code.
Their key sets differ on tool, action type, command class, and observation pattern, so they land in separate blocks.

\paragraph{Articulation crossing: clear phase change.}
\terminal{}, task \texttt{processing-pipeline}.
Two steps immediately before and after an articulation point; the annotator labelled the crossing a \emph{clear phase change}.

\smallskip\noindent\small
\begin{tabularx}{\linewidth}{@{}l@{\;\;}X@{}}
\textbf{Before} & \texttt{read\_file} \texttt{collect\_data.sh} $\to$ shell script body \\[2pt]
\textbf{After} & \texttt{read\_file} \texttt{run\_pipeline.sh} $\to$ orchestration calling \texttt{collect}, \texttt{process}, \texttt{report} \\
\end{tabularx}
\normalsize

\smallskip\noindent
The rollout shifts from examining an individual component to the top-level orchestration wrapper.
The articulation point captures this strategy transition.

\paragraph{Core block: productive progress.}
\mcp{}, task \texttt{mcpbench\_10}.
Three consecutive steps from a block labelled \emph{core} by the diffusion field; the annotator judged the block \emph{productive progress}.

\smallskip\noindent\small
\begin{tabularx}{\linewidth}{@{}l@{\;\;}X@{}}
\textbf{1} & \texttt{MCP\_sum([10.3, 5.4, 8.2])} $\to$ \texttt{23.9} \\[2pt]
\textbf{2} & \texttt{MCP\_sum([52, 29, 43])} $\to$ \texttt{124} \\[2pt]
\textbf{3} & \texttt{MCP\_sum([52.2, 29.1, 43.2])} $\to$ \texttt{124.5} \\
\end{tabularx}
\normalsize

\smallskip\noindent
The rollout makes consistent, goal-directed MCP tool calls.
All observed visitors attain the task-local maximum normalized reward, placing the block firmly in the productive core.

\paragraph{Trap block: low-outcome exploration.}
\swe{}, task \texttt{django-14580}.
Three steps from a block labelled \emph{trap}; the annotator judged the block \emph{exploration without convergence}.

\smallskip\noindent\small
\begin{tabularx}{\linewidth}{@{}l@{\;\;}X@{}}
\textbf{1} & \texttt{view} \texttt{serializer.py:270--280} $\to$ \texttt{TypeSerializer} def \\[2pt]
\textbf{2} & \texttt{view} \texttt{serializer.py:170--190} $\to$ \texttt{functools} imports \\[2pt]
\textbf{3} & \texttt{view} \texttt{serializer.py:400--450} $\to$ (empty---file ends) \\
\end{tabularx}
\normalsize

\smallskip\noindent
The rollout reads scattered sections of the same file without converging on a fix.
Block purity is 0.125 (most visitors fail), placing it in the trap region.
Notably, 36\% of annotated trap blocks are labelled exploration rather than dead-end, consistent with the main-text observation that traps are low-outcome regions, not necessarily useless behavior.

\subsection{Robustness of Supply and Demand Profiles}
\label{app:sensitivity}

\begin{figure*}[h]
\centering
\includegraphics[width=\textwidth]{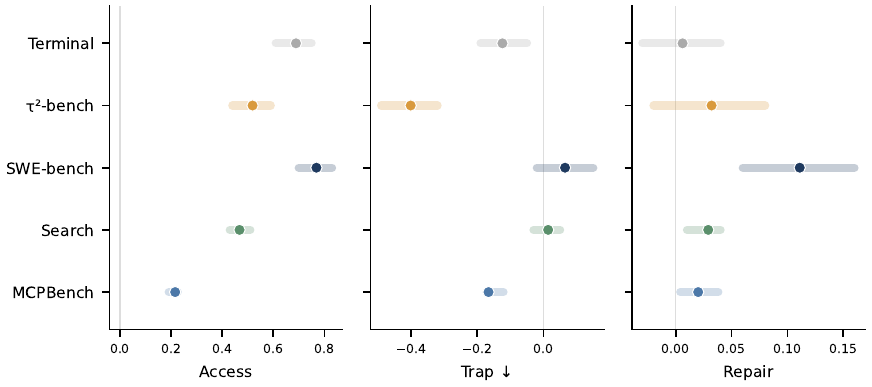}
\caption{Demand-sign robustness across hyperparameter sweeps. Dots show baseline values; shaded bars show the range across core-quantile, trap-quantile, Laplace-smoothing, diffusion, and mutual-$k$NN perturbations. All qualitative signs are stable except the borderline \terminal{} Trap sign under the largest mutual-$k$NN setting.}
\label{fig:robustness}
\end{figure*}

We report task-bootstrap CIs in the main text and additionally sweep the graph-role hyperparameters.
The qualitative signs used in the paper are stable under core-quantile, trap-quantile, Laplace-smoothing, diffusion, and mutual-$k$NN perturbations.
For \mcp{} under the continuous treatment, Access remains positive and Trap remains negative across the same perturbations, while Repair stays small.
Trap-averse conclusions for \mcp{} and \taub{} are stable across trap thresholds, and \swe{} remains the split with the strongest positive Repair demand across the baseline sweep.
The only borderline sign in the sweeps is \terminal{} Trap under the largest $k$NN setting, which is why the main text treats TerminalBench as mildly rather than strongly trap-averse.
As an external replication check, an independently ingested TerminalBench trajectory set reproduces the three-axis demand shape (Access positive, Trap negative, Repair near zero).
Figure~\ref{fig:robustness} shows the baseline demand values alongside the sweep ranges; all signs remain stable except the borderline \terminal{} Trap under the largest $k$NN setting.

\subsection{Full Supply and Demand Tables}
\label{app:supply_demand_tables}

Tables~\ref{tab:model_profiles} and~\ref{tab:demand_vectors} give the exact numerical values underlying Figure~\ref{fig:supply_demand}.
Table~\ref{tab:model_profiles} reports model supply as task-centered residuals with task-bootstrap confidence intervals, while Table~\ref{tab:demand_vectors} reports benchmark demand as resolved-minus-failed contrasts (or the reward-weighted analogue for \mcp{}); positive Access and Repair indicate that high-outcome traffic uses those events more often, whereas negative Trap indicates stronger avoidance of low-outcome regions.

\begin{table*}[h]
\centering
\small
\setlength{\tabcolsep}{4pt}
\begin{tabular}{lccc}
\toprule
\textbf{Model} & \textbf{Access} & \textbf{Trap} & \textbf{Repair} \\
\midrule
DeepSeek-R1 & $-.022\;[-.043,-.001]$ & $+.012\;[-.010,+.035]$ & $+.007\;[-.007,+.020]$ \\
DeepSeek-V3.2 & $+.098\;[+.077,+.118]$ & $+.018\;[-.003,+.038]$ & $+.040\;[+.027,+.052]$ \\
Gemini-2.5-Flash & $-.022\;[-.040,-.004]$ & $-.024\;[-.041,-.006]$ & $-.017\;[-.025,-.009]$ \\
Qwen3-235B & $+.016\;[-.002,+.035]$ & $+.028\;[+.009,+.047]$ & $+.011\;[-.000,+.022]$ \\
Qwen3-Next & $-.092\;[-.116,-.067]$ & $-.042\;[-.067,-.017]$ & $-.049\;[-.058,-.039]$ \\
\bottomrule
\end{tabular}
\caption{Model supply profiles with 2,000 task-bootstrap $95\%$ CIs. Values are task-centered event residuals; because graph-valid task support differs by model, columns need not sum to zero exactly.}
\label{tab:model_profiles}
\end{table*}

\begin{table*}[h]
\centering
\small
\setlength{\tabcolsep}{4pt}
\begin{tabular}{lccc}
\toprule
\textbf{Benchmark} & \textbf{Access} & \textbf{Trap} & \textbf{Repair} \\
\midrule
\mcp{} & $+.216\;[+.182,+.249]$ & $-.166\;[-.199,-.130]$ & $+.020\;[-.004,+.042]$ \\
\searchb{} & $+.468\;[+.423,+.508]$ & $+.014\;[-.026,+.054]$ & $+.029\;[+.007,+.051]$ \\
\swe{} & $+.769\;[+.713,+.822]$ & $+.065\;[-.009,+.148]$ & $+.111\;[+.049,+.175]$ \\
\taub{} & $+.519\;[+.456,+.579]$ & $-.401\;[-.467,-.333]$ & $+.032\;[-.014,+.083]$ \\
\terminal{} & $+.689\;[+.637,+.738]$ & $-.124\;[-.177,-.067]$ & $+.006\;[-.024,+.038]$ \\
\bottomrule
\end{tabular}
\caption{Benchmark demand profiles with 2,000-bootstrap $95\%$ CIs. Negative Trap means high-outcome traffic avoids low-outcome regions.}
\label{tab:demand_vectors}
\end{table*}

\subsection{Additional Trap-Aware Recovery Details}
\label{app:trap_intervention_more}

\paragraph{Recovery arms.}
The three prefix-fork arms defined in \S\ref{sec:prefix_fork} separate two single-factor recovery routes.
\emph{Baseline} continues from the trigger snapshot at the baseline temperature $T{=}0.6$ with no inserted note.
\emph{Hot} switches the continuation to $T{=}0.9$, top-$p{=}0.95$ with no note, isolating the effect of increased sampling diversity.
\emph{Note} appends the diagnosis note at $T{=}0.6$, isolating the effect of evidence-grounded localization.

\begin{table}[h]
\centering
\small
\setlength{\tabcolsep}{5pt}
\begin{tabular}{ll}
\toprule
\textbf{Detector setting} & \textbf{Value} \\
\midrule
Trap similarity threshold & 0.35 \\
Trap--core margin & 0.03 \\
Warmup steps & 2 \\
Cooldown steps & 4 \\
Max triggers per rollout & 1 \\
Allowed intents & edit, submit \\
Observation-key gate & none \\
File-locality gate & require a \texttt{FILE\_PATH:*} cue \\
Rollout step budget & 30 total steps per arm \\
\bottomrule
\end{tabular}
\caption{Fixed detector and continuation settings used in the 500-instance \swe{} recovery study across all three providers.}
\label{tab:detector_hparams}
\end{table}

\paragraph{Conservative diagnosis note.}
The trap-side note is a fixed template populated at fire time.
The operative content is: re-read the failing test, traceback, or last database observation; localize to the smallest directly relevant function or read-only check; make one minimal change consistent with the evidence; and revise or discard the current edit/action if the evidence does not support it.
The slots are populated only from the agent's own per-step log: last command/action, last observation signal, implicated file or target keys, detector confidence, and a six-way family-level trap diagnosis.
No gold patch, verified-test list, or other oracle signal is used.

\paragraph{Tested instances.}
The main \method{}-guided recovery experiment uses the full 500-instance \swe{} Verified pool at seed $11$, covering all 12 \swe{} repositories that appear in the cx-cmu trajectory release: \texttt{django} ($231$), \texttt{sympy} ($75$), \texttt{sphinx-doc} ($44$), \texttt{scikit-learn} ($32$), \texttt{matplotlib} ($34$), \texttt{pytest-dev} ($19$), \texttt{astropy} ($22$), \texttt{pydata} ($22$), \texttt{pylint-dev} ($10$), \texttt{psf} ($8$), \texttt{mwaskom} ($2$), \texttt{pallets} ($1$).
The fired-subset analysis in the upper block of Table~\ref{tab:swe_4arm_intervention} uses each provider's own complete three-arm set ($n{=}196$ for GLM-5.1, $206$ for DeepSeek-V4-Pro, $214$ for Qwen3.6-35B-A3B); the common-fired analysis in the lower block restricts to the $96$ instances on which all three providers fire and produce a complete three-arm set.

\paragraph{Paired contrast statistics.}
Table~\ref{tab:swe_4arm_contrasts} reports the full set of paired contrasts that summarise Table~\ref{tab:swe_4arm_intervention}; the main-text intervention $p$-values refer to these same tests.
All confidence intervals use the cluster bootstrap with $10{,}000$ resamples clustered by instance id; $p$-values are one-sided paired-sign permutation against the alternative $\Delta > 0$.

\begin{table*}[h]
\centering
\small
\setlength{\tabcolsep}{5pt}
\renewcommand{\arraystretch}{1.05}
\begin{tabular}{llccc}
\toprule
\textbf{Subset / pool} & \textbf{Contrast} & \textbf{$\Delta$ (pp)} & \textbf{95\% CI (pp)} & \textbf{$p$} \\
\midrule
\multirow{2}{*}{GLM-5.1 ($n{=}196$)}        & Hot $-$ Baseline       & $+4.59$ & $[+0.0,+9.2]$  & $0.037$\,\checkmark \\
                                            & Note $-$ Baseline & $+1.53$ & $[-3.6,+6.6]$  & $0.348$ \\
\midrule
\multirow{2}{*}{DeepSeek-V4-Pro ($n{=}206$)}& Hot $-$ Baseline       & $+3.40$ & $[-1.5,+8.7]$  & $0.132$ \\
                                            & Note $-$ Baseline & $+0.97$ & $[-4.4,+6.3]$  & $0.428$ \\
\midrule
\multirow{2}{*}{Qwen3.6-35B-A3B ($n{=}214$)}& Hot $-$ Baseline       & $+1.40$ & $[-2.3,+5.1]$  & $0.332$ \\
                                            & Note $-$ Baseline & $+4.67$ & $[+0.0,+9.3]$ & $0.036$\,\checkmark \\
\midrule
\multirow{2}{*}{Per-provider fired pooled ($n{=}616$ cells)} & Hot $-$ Baseline       & $+3.08$ & $[+0.3,+5.9]$ & $0.016$\,\checkmark \\
                                            & Note $-$ Baseline & $+2.44$ & $[-0.6,+5.5]$  & $0.064^{\circ}$ \\
\midrule
\multirow{2}{*}{Common-fired pooled ($n{=}288$ cells)} & Hot $-$ Baseline       & $+3.82$ & $[+0.0,+8.0]$  & $0.045$\,\checkmark \\
                                            & Note $-$ Baseline & $+2.78$ & $[-2.1,+7.6]$  & $0.148$ \\
\bottomrule
\end{tabular}
\caption{Paired contrasts for Table~\ref{tab:swe_4arm_intervention}. Per-provider rows use each provider's own fired iids; the two pooled rows stack provider $\times$ iid cells (clustered by iid). \checkmark\,$p{<}0.05$; $^{\circ}$\,$p{<}0.10$.}
\label{tab:swe_4arm_contrasts}
\end{table*}

\subsection{Trap-Aware Intervention Case Studies}
\label{app:trap_case_studies}

\paragraph{Illustrative intervention cases.}
The following three cases separate a positive xarray example, a positive Django example, and a regression-style failure mode.

\paragraph{Case 1: \texttt{pydata\_\allowbreak{}\_xarray-4094}.}
The same trigger snapshot drives three qualitatively different recovery patterns across providers (Table~\ref{tab:case_xarray_4094}).
The detector fires while the agent is already reading \texttt{xarray/core/\allowbreak{}dataarray.py}, with trigger keys \texttt{ACTION:edit}, \texttt{CMD:sed}, \texttt{FILE\_PATH:xarray/core/\allowbreak{}dataarray.py}, and an earlier \texttt{ValueError} reproduction context.
Across providers, the raw assistant traces converge on the same local diagnosis: after selecting one variable from the stacked array, the stacked coordinate remains as a scalar coordinate, so later merging the per-variable arrays back into a dataset creates a coordinate conflict.
The divergence appears one refinement later.
Qwen3.6-35B-A3B under \emph{Baseline} and \emph{Hot} spends the post-trigger budget on reproduction and environment repair (missing \texttt{numpy}, then a \texttt{numpy} version mismatch) and never submits a patch.
Under \emph{Note}, it eventually reaches the right local fix idea and edits the key line to drop the stacked coordinate after \texttt{squeeze(drop=True)}, but the first version uses \texttt{drop\_vars(dim)} and then fails on the broader multi-dimensional round-trip test because the coordinate is not always present.
DeepSeek-V4-Pro shows a milder version of the same phenomenon: \emph{Baseline} reaches the right diagnosis but commits the brittle \texttt{drop(dim)} variant, whereas either \emph{Hot} or \emph{Note} is enough to move it to the safer \texttt{drop\_vars(dim,\allowbreak{} errors="ignore")} form.
GLM-5.1 is the easiest case: once triggered, both non-Baseline arms jump directly to that safe one-line fix.

\begin{table}[h]
\centering
\scriptsize
\setlength{\tabcolsep}{4pt}
\begin{tabular}{lccc}
\toprule
\textbf{Provider} & \textbf{Baseline} & \textbf{Hot} & \textbf{Note} \\
\midrule
GLM-5.1            & no patch          & \cellcolor{bestcolor}\checkmark\,(572\,ch) & \cellcolor{bestcolor}\checkmark\,(572\,ch) \\
DeepSeek-V4-Pro    & \texttimes (550\,ch) & \cellcolor{bestcolor}\checkmark\,(572\,ch) & \cellcolor{bestcolor}\checkmark\,(572\,ch) \\
Qwen3.6-35B-A3B    & no patch          & no patch & \texttimes (555\,ch) \\
\bottomrule
\end{tabular}
\caption{\texttt{pydata\_\allowbreak{}\_xarray-4094} per-arm outcomes. \checkmark\ marks resolved; \texttimes\ marks a submitted patch that fails the verified tests; ``no patch'' marks an empty submission. Parentheses show patch length in characters.}
\label{tab:case_xarray_4094}
\end{table}

\paragraph{Case 2: \texttt{django\_\allowbreak{}\_django-13128}.}
The trigger fires later in the rollout, after the agent has already narrowed the bug to \texttt{django/db/models/\allowbreak{}expressions.py} and is inspecting \texttt{CombinedExpression} with an \texttt{OBS:error} context (Table~\ref{tab:case_django_13128}).
The trap here is not a wrong first diagnosis but a low-yield validation loop.
In the GLM-5.1 \emph{Baseline} trace, the model states the correct diagnosis almost immediately in its own scratch reasoning: subtracting two temporal fields is being inferred as \texttt{DateTimeField} when it should produce \texttt{DurationField}.
But the baseline rollout then burns budget on environment setup and brittle ad-hoc probes, repeatedly re-deriving the same explanation without committing a final patch.
\emph{Hot} and \emph{Note} both shorten that loop.
The successful GLM arms add a narrow \texttt{\_resolve\_\allowbreak{}output\_field()} override directly to \texttt{CombinedExpression}, then fix one subtle bug exposed by a direct probe: comparing field \emph{instances} with \texttt{==} is too strict, so the final patch compares field types / internal types and returns \texttt{fields.\allowbreak{}DurationField()} only for temporal subtraction.
The \emph{Note} arm is especially illustrative: after trigger it stays anchored to the smallest relevant region---\texttt{CombinedExpression} and the temporal-subtraction tests---rather than relaxing the global mixed-type logic elsewhere in \texttt{BaseExpression}.
DeepSeek-V4-Pro shows the complementary provider-specific pattern on the same instance.
Only \emph{Hot} reaches a submitted fix; the \emph{Note} arm remains in the expression-tree / test-harness analysis loop and ends with an empty submission, matching the aggregate result that this provider benefits mainly from extra sampling diversity rather than the diagnosis note itself.

\begin{table}[h]
\centering
\scriptsize
\setlength{\tabcolsep}{4pt}
\begin{tabular}{lccc}
\toprule
\textbf{Provider} & \textbf{Baseline} & \textbf{Hot} & \textbf{Note} \\
\midrule
GLM-5.1         & no patch & \cellcolor{bestcolor}\checkmark\,(1057\,ch) & \cellcolor{bestcolor}\checkmark\,(1001\,ch) \\
DeepSeek-V4-Pro & no patch & \cellcolor{bestcolor}\checkmark\,(1175\,ch) & no patch \\
\bottomrule
\end{tabular}
\caption{\texttt{django\_\allowbreak{}\_django-13128} per-arm outcomes for the two providers discussed in the text. Parentheses show patch length in characters.}
\label{tab:case_django_13128}
\end{table}

\paragraph{Case 3: failure mode on \texttt{django\_\allowbreak{}\_django-14311}.}
The same framework can also hurt performance when the note over-focuses the continuation on an overly narrow local pattern.
On \texttt{django\_\allowbreak{}\_django-14311}, the trigger fires while GLM-5.1 is already inspecting \texttt{tests/utils\_tests/\allowbreak{}test\_autoreload.py}, with trigger keys anchored to an edit candidate in the autoreload path and an earlier \texttt{RuntimeError} context.
All three arms correctly identify the broad bug family: \texttt{get\_child\_\allowbreak{}arguments()} should reconstruct the \texttt{-m} target from \texttt{\_\_\allowbreak{}main\_\_.\allowbreak{}\_\_spec\_\_}.
But the Note arm overcommits to the immediately visible package-with-\texttt{\_\_\allowbreak{}main\_\_} example from \texttt{utils\_tests.\allowbreak{}test\_module}.
It rewrites the logic to always use \texttt{\_\_\allowbreak{}spec\_\_.name.\allowbreak{}removesuffix('.\_\_\allowbreak{}main\_\_')}, which fixes the package case but breaks the non-package module case and regresses three previously passing tests.
By contrast, the successful \emph{Baseline} arm makes the minimal string replacement \texttt{name.\allowbreak{}replace('.\_\_\allowbreak{}main\_\_', '')}, and the successful \emph{Hot} arm reaches the same fix with a slightly different edit path.
This is a concrete example of a trap-aware note becoming too prescriptive: it improves localization, but can also bias the model toward an over-specialized repair if the visible trigger context is only one branch of the full behavioral contract.

\begin{table}[h]
\centering
\scriptsize
\setlength{\tabcolsep}{4pt}
\begin{tabular}{lccc}
\toprule
\textbf{Provider} & \textbf{Baseline} & \textbf{Hot} & \textbf{Note} \\
\midrule
GLM-5.1 & \checkmark\,(694\,ch) & \checkmark\,(953\,ch) & \texttimes\,(852\,ch) \\
\bottomrule
\end{tabular}
\caption{A regression-style intervention case. On \texttt{django\_\allowbreak{}\_django-14311}, the Note arm fails \texttt{test\_run\_as\_non\_django\_\allowbreak{}module\_non\_package} and three pass-to-pass autoreload tests, while the other two arms resolve the instance.}
\label{tab:case_failure_django_14311}
\end{table}

\subsection{Agent Framework and Recovery Templates}
\label{app:templates}

The \swe{} continuation agent is a single-tool bash agent.
We reproduce here the templates that fully specify what the agent sees on every step and what each arm injects: the system prompt that fixes the response format, the issue-statement user message, the observation wrapper used after every command, and the 5-block diagnosis-note template that the \emph{Note} arm injects at the trigger step (with the \emph{Baseline} and \emph{Hot} arms injecting nothing).

\paragraph{System prompt.}
Used as the first message of every rollout, on every arm.

\begin{tcolorbox}[colback=gray!10, colframe=gray!50, boxrule=0.5pt, breakable]
\tiny
\begin{Verbatim}[breaklines,breakanywhere]
You are a software engineer tasked with solving a GitHub issue. You have access to a bash shell to explore the repository, understand the codebase, locate the relevant code, and make the necessary changes to fix the issue.

## Response Format

Every response must contain exactly TWO sections:

1. **THOUGHT**: Your reasoning about what to do next. Analyze the problem, plan your approach, and explain your next step.

2. **ACTION**: A single bash command to execute. This must be wrapped in a bash code block.

Example response:

THOUGHT:
I need to find the file that contains the buggy function.

ACTION:
```bash
find . -type f -name "*.py" | xargs grep -l "def process_data"
```

## Important Rules

- Each response must contain exactly ONE bash command in the ACTION section.
- To submit your solution when done, use the special command:
  ```bash
  submit
  ```
- Do NOT run interactive commands (vim, nano, python REPL, etc.).
- Do NOT use `git push` or modify remote state.
- Keep your changes minimal and focused on the issue.
- Always verify your fix with relevant tests before submitting.
- If you need to edit a file, use `sed`, `awk`, or heredoc redirects.
- If a command produces too much output, pipe through `head` or `tail`.
\end{Verbatim}
\end{tcolorbox}

\paragraph{Initial user message.}
Inserted once at step $0$ to communicate the GitHub issue. The placeholder \verb|{problem_statement}| is filled with the verbatim \swe{} Verified \texttt{problem\_statement} field.

\begin{tcolorbox}[colback=gray!10, colframe=gray!50, boxrule=0.5pt, breakable]
\tiny
\begin{Verbatim}[breaklines,breakanywhere]
Here is the GitHub issue to solve:

<issue>
{problem_statement}
</issue>

You are in the repository root. The repository has already been checked out to the correct commit. Explore the repo, understand the issue, make the fix, and run tests to verify. When you are confident your fix is correct, use the `submit` command.
\end{Verbatim}
\end{tcolorbox}

\paragraph{Per-step observation wrapper.}
After every \texttt{ACTION} bash command, the harness executes it inside the docker workspace and returns the (truncated) combined stdout/stderr to the agent as a user message wrapped in:

\begin{tcolorbox}[colback=gray!10, colframe=gray!50, boxrule=0.5pt]
\tiny
\begin{Verbatim}[breaklines,breakanywhere]
OBSERVATION:
{observation}
\end{Verbatim}
\end{tcolorbox}

\paragraph{Trap-diagnosis note (Note arm).}
At the trigger step both \textsc{Note}-on arms inject the following message as a user message immediately after the last observation, before the agent's next \texttt{THOUGHT}/\texttt{ACTION} turn. The header marks the message as internal; the five slots \verb|{ctx_command}|, \verb|{ctx_signal}|, \verb|{ctx_files}|, \verb|{ctx_confidence}|, \verb|{ctx_trap_pattern}| are populated only from the agent's own per-step log and a per-pattern diagnosis sidecar; no gold-patch or verified-test information is used.

\begin{tcolorbox}[colback=gray!10, colframe=gray!50, boxrule=0.5pt, breakable]
\tiny
\begin{Verbatim}[breaklines,breakanywhere]
[INTERNAL DIAGNOSTIC -- not visible to graders]
A trajectory pattern previously associated with low task-resolution rates fired at this step. Concretely:

  - Last command: {ctx_command}
  - Last error/test signal: {ctx_signal}
  - File(s) touched in this region: {ctx_files}
  - Detector confidence (trap similarity): {ctx_confidence}

  Trap pattern (from prior failed trajectories with this signature):
  {ctx_trap_pattern}

Before continuing:
  1. Re-read the failing test/traceback for the specific assertion or unexpected value (do not rely on memory of earlier steps).
  2. Localize to the smallest function or class implicated by that evidence, in the file(s) above.
  3. Propose ONE minimal change consistent with the evidence; do not rewrite unrelated code.
  4. Run the narrowest relevant test or check before submitting.
  5. If your current patch is not supported by the error/test evidence, revise or discard it.

Respond in the normal THOUGHT / ACTION format with exactly one bash command.
\end{Verbatim}
\end{tcolorbox}

\paragraph{Slot filling.}
At trigger time the slots are populated from the agent's own per-step record (no oracle signal):
\verb|{ctx_command}| is the most recent \texttt{ACTION} bash command, truncated to $200$ characters;
\verb|{ctx_signal}| is the most recent observation passed through a fixed regular-expression family for traceback / pytest / syntax / import / missing-file / timeout cues, truncated to $300$ characters;
\verb|{ctx_files}| is the top-$3$ file paths from the trigger key set (only files the agent itself has opened in its own commands);
\verb|{ctx_confidence}| is the trap similarity score $\mathrm{sim}_{\mathrm{trap}}$ formatted to two decimals;
\verb|{ctx_trap_pattern}| is the family-level diagnosis string for the matched trap pattern (one of six pre-specified families), read from a sidecar JSON keyed on the trap key-set fingerprint.
The \emph{Baseline} and \emph{Hot} arms inject nothing at the trigger step; the conversation proceeds directly from the trigger observation to the agent's next \texttt{THOUGHT}/\texttt{ACTION} turn at the arm's continuation temperature.

\section{Algorithmic Details}
\label{app:algos}

\subsection{Signature Key Alphabet}
\label{app:keys}

A step's symbolic key set $K_i$ contains keys from the channels in Table~\ref{tab:keys}.
Keys are case-sensitive and prefix-tagged, so different channels never collide in the IDF dictionary.
Observation keys are produced by fixed regular-expression families over the first $2{,}000$ characters of the tool response; the families include error, traceback, test pass/fail, syntax error, import error, missing file, permission error, timeout, and explicit success messages.
The search-specific keys (\texttt{URL\_DOMAIN}, \texttt{QUERY\_NOVELTY}, \texttt{EVIDENCE\_COUNT}) are pre-specified from the search tool-call/result format, not selected from outcome inspection.

\begin{table}[h]
\centering
\scriptsize
\setlength{\tabcolsep}{3pt}
\begin{tabularx}{\linewidth}{lX}
\toprule
\textbf{Channel} & \textbf{Information captured} \\
\midrule
Tool name & Tool or API name after benchmark-specific prefix stripping. \\
Action type & Coarse action class such as bash, edit, read, search, submit, or other. \\
Command class & Shell-command families such as pytest, grep, cat, sed, python, and git. \\
Observation pattern & Error, pass/fail, timeout, missing-file, and success cues. \\
File cue & Last path components and file extensions mentioned in the action. \\
Temporal phase & Early, middle, or late third of the rollout. \\
Search-only & URL/result domain (eTLD$+1$), query novelty (new/refine/repeat), and cumulative distinct-domain count buckets. \\
\bottomrule
\end{tabularx}
\caption{Compact signature key alphabet used by the signature extractor.}
\label{tab:keys}
\end{table}

\subsection{Reward Field, Core and Trap Masks, Block Quotient Graph}
\label{app:roles_math}

The \emph{block quotient graph} $G_B=(B,E_B)$ has one node per non-trivial BCC and an undirected edge $(b,b')\in E_B$ weighted by the number of slices shared between the two blocks.
Edge weights are row-stochastically normalized into a transition matrix $P_B$, and the teleport-$\alpha$ diffusion update is
\begin{equation}
v^{(t+1)} = \alpha\, P_B^{\top} v^{(t)} + (1-\alpha)\, s,
\label{eq:diffuse}
\end{equation}
where $\mathcal{R}_b$ is the set of rollouts visiting block $b$, and the seed $s_b=\hat\mu_b-\bar y_t$ is the Laplace-smoothed excess outcome of block $b$ relative to the task average, with $\hat\mu_b=(\sum_{r\in\mathcal{R}_b} y_r+\tfrac{1}{2})/(|\mathcal{R}_b|+1)$.
We iterate Eq.~\ref{eq:diffuse} for $T=24$ steps with $\alpha=0.65$ and L$\infty$-normalize to $[-1,1]$.
Let $Q^+_{.75}$ be the 75th percentile of positive field values and $Q^-_{.25}$ the 25th percentile of negative field values.
The role overlays used in the main text are then:
\begin{align}
C_t &= \{b:\, v^{(T)}_b>0 \;\wedge\; v^{(T)}_b\geq Q^+_{.75}\}, \\
B_t &= \{b:\, v^{(T)}_b<0 \;\wedge\; v^{(T)}_b\leq Q^-_{.25}\}.
\end{align}
Note that $B_t$ is a direct bottom-quartile mask of the negative diffusion field, which avoids a circularity between the trap definition and the $R_r$ event.
Articulation points remain useful for graph visualization and the RQ1 sanity annotation, but they are not used as a main rollout event or as a recovery target.

\end{document}